\documentclass{article}
\usepackage[preprint]{log_2024}			

\usepackage{booktabs}						
\usepackage{multirow}						
\usepackage{amsfonts}						
\usepackage{graphicx}						
\usepackage{duckuments}						

\usepackage{makecell}

\usepackage[numbers,compress,sort]{natbib}	

\title[Graph Neural Networks on Graph Databases]{Graph Neural Networks on Graph Databases}

\author[D. Lopushanskyy et al.]{%
Dmytro Lopushanskyy\\
University of Oxford\\
\email{dmytro.lopushanskyy@keble.ox.ac.uk}\And
Borun Shi\\
Neo4j\\
\email{brian.shi@neo4j.com}
}

\usepackage{xcolor}

\usepackage{ifxetex}
\usepackage{ifluatex}
\newif\ifxetexorluatex 
\ifnum 0\ifxetex 1\fi\ifluatex 1\fi>0
   \xetexorluatextrue
\fi

\ifxetexorluatex
  \usepackage{fontspec}
\else
  \usepackage[T1]{fontenc}
  \usepackage[utf8]{inputenc}
\fi

\usepackage{textcomp}
\usepackage{xcolor}
\usepackage{listings}
\usepackage{upquote}

\definecolor{keyword}{HTML}{2771a3}
\definecolor{pattern}{HTML}{b53c2f}
\definecolor{string}{HTML}{be681c}
\definecolor{relation}{HTML}{7e4894}
\definecolor{variable}{HTML}{107762}
\definecolor{comment}{HTML}{8d9094}

\lstdefinelanguage{cypher}
{
	morekeywords={
		MATCH, OPTIONAL, WHERE, NOT, AND, OR, XOR, RETURN, DISTINCT, ORDER, BY, ASC, ASCENDING, DESC, DESCENDING, UNWIND, AS, UNION, WITH, ALL, CREATE, DELETE, DETACH, REMOVE, SET, MERGE, SET, SKIP, LIMIT, IN, CASE, WHEN, THEN, ELSE, END,
		INDEX, DROP, UNIQUE, CONSTRAINT, EXPLAIN, PROFILE, START,
            SELECT, FROM, JOIN, ON
	}
}

\newcommand{\mycdots}{\cdot\!\cdot\!\cdot}
\begin{document}

\maketitle

\begin{abstract}
Training graph neural networks on large datasets has long been a challenge. Traditional approaches include efficiently representing the whole graph in-memory, designing parameter efficient and sampling-based models, and graph partitioning in a distributed setup. 
Separately, graph databases with native graph storage and query engines have been developed, which enable time and resource efficient graph analytics workloads.
We show how to directly train a GNN on a graph DB, by retrieving minimal data into memory and sampling using the query engine. Our experiments show resource advantages for single-machine and distributed training. Our approach opens up a new way of scaling GNNs as well as a new application area for graph DBs.
\end{abstract}

\section{Introduction}
Graph-structured data is ubiquitous, underlying many of the complex systems we interact with daily, from social and transportation networks to molecular structures and knowledge graphs. As these datasets continue to grow in volume and complexity, there is an increasing need for machine learning systems that can identify meaningful patterns within graphs. Graph neural networks (GNNs) have emerged as a state-of-the-art solution for many of these graph machine learning tasks, driving advances in fields as diverse as recommender systems, fraud detection, and drug discovery.

Despite their success, GNNs still face significant challenges when it comes to scalability. Large real-world graphs often contain billions of nodes and edges, bringing an additional memory bottleneck. There exist several prior approaches that improve scalability. Compressing a sparse graph's in-memory representation, for example, in Compressed Sparse Row (CSR)\cite{csr}, reduces memory needed to store edges. Parameter sharing techniques and sampling-based GNNs\cite{graphsage,chen2018fastgcn, zou2019layer} which reduce neighbourhood explosion improve training speed and hence the applicability of models to large graphs.
Distributed GNN training frameworks \cite{zheng2020distdgl, dist-PyTorch} enable one to scale training speed with respect to hardware resources available.

However, the scalability issues of GNNs have not been fully resolved. In-memory representation of large, heterogeneous and feature-rich graphs remains memory-bound. In distributed setups, new trade-offs arise between memory efficiency and computational speed, which are caused by data locality issues and inter-node communication overhead brought by graph partitioning. There are also other practical issues facing GNNs.
Challenges in handling real-world dynamic graphs include linear time graph update (in CSR), ensuring data consistency, and synchronisation for partitioned graphs. The Extract, Transform and Load (ETL) layer adopted by graph ML remains inefficient, as it frequently involves the update and reloading large CSVs.

Separately, graph databases, query languages and query engines have evolved. 
A graph DB treats nodes and edges as primary entities in its graph schema, storage format, query language, as well as in the operators implemented by the query engine. This allows time- and space-efficient execution of native graph queries. 
For example, finding all nodes reachable from a given source node can be expressed and executed conveniently in a graph DB. If the data is instead modelled as tables and queried through SQLs, complex many-to-many joins are involved. Typical database properties that exist in relational databases, such as ACID transactions, online transactional processing (OLTP) and online analytical processing (OLAP) workloads, are supported by graph DBs. Naturally, it is increasingly common to store large graphs in various graph DBs within the industry.

The lack of synergy between graph ML and DB is largely due to the historical fact that they started as separate fields. However, it becomes increasingly obvious that a tighter coupling improves graph ML pipelines. To start with, data copy from the DB into GNN workload is wasteful and slow. On a deeper level, both subjects are often solving the same underlying problems on graphs, which include efficient representation and retrieval of graph information. 

We leverage graph DBs as a core component of GNN. We show how to formulate graph ML tasks like neighbour sampling and feature retrieval as queries for graph DBs. As a result, we significantly improve the memory efficiency of training GNNs.

\section{Background}

\subsection{Graph neural networks}
One of the dominant models for learning on graphs is message passing neural networks (MPNNs), a class of GNNs.  An MPNN operates on all the nodes in a graph, for each node recursively collects information from neighbouring nodes, and aggregates and updates the node's information until convergence. Given a graph $G = (V, E)$ with node initial features $x_v^{(0)} \in \mathbb{R}^{1 \times n}$ for $u \in V$, an MPNN updates a node embedding $x_v^k$ at layer $k$ by the equation:
\begin{equation}
\mathbf{x}_v^{(k)} = \phi^{(k)} \left( \mathbf{x}_v^{(k-1)}, \bigoplus_{u \in \mathcal{N}(v)} \psi^{(k)} \left( \mathbf{x}_v^{(k-1)}, \mathbf{x}_u^{(k-1)}\right) \right),
\label{eq:mpnn}
\end{equation}
where $\psi$ is a message function, $\bigoplus$ is an injective multiset function, and $\phi$ is a update function, and $\mathcal{N}(v)$ is some neighbourhood of $u$ such as $\{u \in V : (v, u) \in E\}$.

Reducing the number of parameters in $\phi$, $\bigoplus$, $\psi$ improves model scalability. The sampling of neighbourhood $\mathcal{N}(v)$ further reduces the worst-case parametric complexity from $O(|V|)$ to some configurable constant $O(S_k)$ for any layer $k$. These methods underpin the majority of modern GNNs, with a good example being (mean-aggregate) GraphSAGE\cite{graphsage}:
\begin{equation}
x_v^{(k)} = \sigma \left( W^{(k)} \cdot \text{CONCAT}\left( x_v^{(k-1)}, \text{MEAN}\left( \{\!\!\{x_u^{(k-1)} \, | \, u \in \mathcal{N}^k(v)\}\!\!\} \right) \right) \right)
\label{eq:graphsage}
\end{equation}
Here $\sigma$ is some non-linear activation function, $W^{(k)}$ is the only layerwise-learnable weight, and $\mathcal{N}^k(v)$ is some layerwise sampled neighbourhood of $v$.

\subsection{Implementation of GNNs}
\label{subsec:background-training-architecture}
There exist several software libraries for GNNs, such as PyTorch Geometric (PyG) \cite{pyg}, the Deep Graph Library (DGL) \cite{wang2019dgl} and TensorflowGNN \cite{tfgnn}. A typical workflow is to load the entire graph from some files (CSVs) into memory for single-machine training.

Distributed training techniques have been developed to speed up the training process by scaling it with respect to the available hardware resources. The starting point is some graph partitioning algorithm, such as METIS \cite{metis, metis-software}. Such algorithms usually achieve a trade-off between good workload-balance to ensure all training processes are well saturated and min-cut across partitions to reduce inter-process communications \cite{cofree-gnn}. Libraries such as DistDGL \cite{zheng2020distdgl} and PyTorch Distributed Data Parallel (DDP) \cite{dist-PyTorch} support such setups. \cite{g3, bytegnn} addresses the engineering bottlenecks in workload balancing, scheduling of pre-sampling steps and CPU utilisations.

\subsection{Graph databases and graph query languages}
There are multiple ways to model graphs within graph DBs. Two of the most common ones are Labelled Property Graph (LPG)\cite{angles2017foundationsmodernquerylanguages} and Resource Description Framework (RDF)\cite{rdf}, each equipped with its own way of querying the graph.

\begin{figure}
    \centering
    \includegraphics[width=0.9\linewidth]{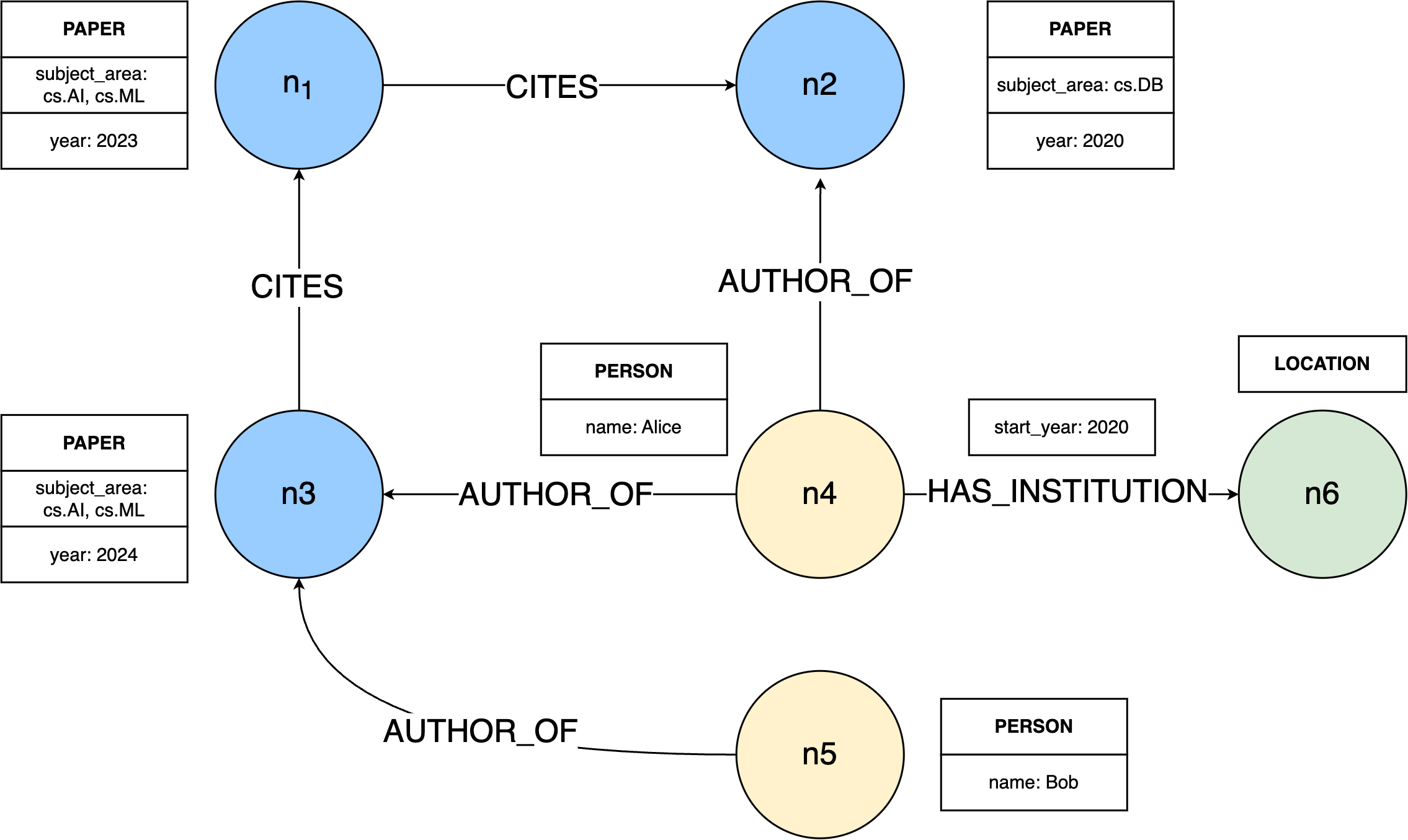}
    \caption{An example labelled property graph. Simple citation graphs such as cora\cite{cora} or ogbn-papers100M\cite{hu2020open} can be modelled by such schema.}
    \label{fig:LPG}
\end{figure}

An LPG is defined as a tuple $(V, E, L, l_v, l_e, K, W, {p_v}, {p_e})$. $V$ denotes the the set of vertices, $E$ the set of edges, $L$ the set of labels, $l_v: V \rightarrow \mathcal{P}(V)$ the label assignment function on vertices, $l_e: E \rightarrow \mathcal{P}(E)$ the label assignment function on edges, $K$ the set of keys (attribute or property names), $W$ the set of values (attribute or property values), $p_v: V \rightarrow K \times W$ the property key-value assignment functions on vertices, $p_e: E \rightarrow K \times W$ the property key-value assignment functions on edges. Loosely speaking, LPG allows the most flexible modelling of any graph. For example, vertices and edges can share the same labels, property keys, and values. Figure~\ref{fig:LPG} shows an example of an LPG graph.

There exist several query languages for LPG, such as Cypher\cite{cypher, cyphersemantics} and openCypher \cite{opencypher}, which form the basis of the recent ISO GQL standard \cite{gql,gqlpaper}. The power of Cypher comes with graph pattern matching and the fact that nodes and edges are treated as first-class entities in LPGs. Figure~\ref{fig:cypherSQL} shows an example Cypher query. 

An increasing number of industrial LPG graph databases support Cypher, such as Neo4j \cite{neo4j}, ArangoDB \cite{arangoDB}, TigerGraph \cite{tigergraph}.

\begin{figure}
 \centering
 \begin{minipage}{0.47\textwidth}
 \captionsetup{type=listing}
 \begin{lstlisting}[language=cypher]
    MATCH (n:PAPER)-[:CITES]->(m:PAPER)
    RETURN DISTINCT m
 \end{lstlisting}
 \end{minipage}
 \hfill
 \begin{minipage}{0.47\textwidth}
 \captionsetup{type=listing}
 \begin{lstlisting}[language=cypher]
    SELECT DISTINCT p.*
    FROM PAPERS p
    JOIN CITES c ON p.id = c.cited_paper_id
 \end{lstlisting}
 \end{minipage}
 \caption{Left: A Cypher query that returns all papers that are cited by at least one other paper. Right: An equivalent SQL query, assuming a reasonable table schema, such as one table PAPERS with properties columns and a second table CITES with columns citing\_paper\_id and cited\_paper\_id.}
 \label{fig:cypherSQL}
 
\end{figure}

RDF, which originated from the semantic web, adopts a different graph data model. The central entities in RDF are triples, usually denoted as $(s, p, o)$, for subject-predicate-object. Therefore, RDF graph databases are also commonly referred to as triple stores. SPARQL\cite{sparql} is a common RDF query language. Implementations of triple stores include RDFox \cite{rdfox}. While it is common to use Cypher over LPGs and SPARQL over RDFs, there exist implementations that allow one to execute one type of query language over another graph data model, such as AWS Neptune \cite{neptune}. With minor modifications, Cypher queries can be translated into SPARQL.

\section{Related Work}
To the best of our knowledge, there is no previous work that trains a GNN directly on a graph database leveraging its query engine. There are several preliminary work that attempts to integrate graph DBs with ML workflows.
K\`uzu \cite{chen2023kzu} extended PyG's default \texttt{featureStore} and \texttt{graphStore}. Being an embedded graph DBMS, the entire graph is loaded into the memory of one machine and converted (in a zero-copy manner) to torch tensors, after which existing single-machine PyG training can be used.

LPG2vec \cite{lpg2vec} and RDF2vec \cite{rdf2vec} are shallow embedding methods that apply directly to the aforementioned graph representations, which can be seen as different heterogeneous extensions of node2vec \cite{node2vec}. \cite{ren2023neural} proposed extending (graph) query engines to handle algebraic operations on these embeddings. Relational deep learning\cite{kumo} applies GNNs on tabular data by first forming a (hyper)graph.

As ML systems evolve and data grows in size, compound systems become increasingly common \cite{compound-ai-blog} as opposed to single monolithic models that are often small. For example, databases recently appeared to be an important retrieval component in the retrieval augmented generation (RAG)\cite{lewis2020retrieval, graphrag, gnnrag} framework for large language models. Our work could be viewed as introducing a more tightly coupled retrieval component within the graph ML pipeline.

While we focus on database support for GNNs, in-database learning \cite{Klbe2023ExplorationOA, khamis2017learning, olteanu2020relational} studies the use of relational databases to support the implementation of standard ML models, such as linear regression.

\section{Training GNN on graph DB}
We now present our approach, which offloads all retrieval steps of a typical GNN training pipeline to the graph DB. In particular, we only retrieve necessary metadata at the start and do not materialise any graph information. For each batch of nodes, we sample their neighbours and retrieve their features through executing queries.
Our approach applies to a variety of GNNs, tasks, datasets and software libraries.

\paragraph{Initial Data Retrieval}
The first step in training GNNs in existing workflows involves loading the graph, with initial features, into memory. Various representation formats, such as CSR, Coordinate Format (COO) and edge list, offer different memory and complexity tradeoffs. Two sets of configurations, one being user-defined hyperparameters, such as the number of layers, and the other being graph metadata, such as the number of nodes and initial feature dimensions, are used to initialise the dataloader and model. 

Up until this point, the graph does not need to be fully materialised. Instead, we only retrieve the necessary graph metadata from the graph in the DB. Figure~\ref{fig:node-metadata} shows the query that retrieves node metadata. Any edge metadata can be similarly obtained by another Cypher query which we include in Appendix~\ref{appendix:edge-metadata}.

For example, when training ogbn-papers100M\cite{hu2020open}, our setup requires less than 1KB of RAM, instead of 48GB in the regular setup, during the initialisation stage.

\begin{figure}
 \centering
 \begin{lstlisting}[language=cypher]
    MATCH (n)
    WITH DISTINCT labels(n) AS NodeTypes, n
    UNWIND NodeTypes AS NodeType
    WITH NodeType, collect(distinct keys(n)) AS AllKeys
    RETURN NodeType, reduce(s = [], k IN AllKeys | s + k) AS Attributes
 \end{lstlisting}
 \caption{Cypher query that returns all relevant metadata for all nodes.}
 \label{fig:node-metadata}
\end{figure}

\paragraph{Neighbour Sampling}
Once the system is initialised with the necessary metadata, it proceeds with the mini-batch training steps. As captured by the update Equation \ref{eq:graphsage}, the key operations are sampling and feature extraction. Typically, the sampling operation has access to the full graph structure and is implemented in-memory. We offload the operation to the database. We construct a query template, as shown in Figure \ref{fig:sampling-query}, which, when supplied with seed nodes and sampling parameters, gets executed on the database. As a result, only the absolute minimal information is materialised in the training process. 

Since there is no keyword in Cypher that is equivalent to \texttt{SAMPLE}, it is not immediately obvious how sampling can be done as part of the query. From a database point of view, sampling does not traditionally fit into either an OLTP or an OLAP workload. One might attempt to use a query similar to Figure\ref{fig:cypherSQL}, and then \texttt{LIMIT} the output to the number of samples. However, the semantics of \texttt{MATCH \dots RETURN \dots} do not have any ordering guarantee. 

A single execution of the query with a specific set of seed nodes on a fresh database often returns results in nondeterministic ordering. On the other hand, repeated executions give the same ordering because of the caching mechanisms implemented by the majority of databases. Therefore, we utilise the combination of an in-line function \texttt{rand()}, \texttt{ORDER BY} and \texttt{LIMIT} to guarantee uniformly randomly sampled neighbourhoods.

Multi-hop neighbourhoods are frequently used to allow nodes to aggregate information from farther away. One approach we could take is to separately query for each hop. Sampled neighbours returned are used as seed nodes for the next query (second hop), and so on. This query is shown in Appendix~\ref{appendix:one-hop-sampling}. However, this increases latency and the resources required. More importantly, it eliminates the advantage of a query engine that is designed to execute multi-hop queries. Hence, our query performs multi-hop sampling by chaining together multi-hop pattern matching.

Node-wise uniform random sampling per layer, as implemented by the original GraphSAGE, requires a \texttt{LIMIT} clause at each hop. Consequently, the query engine has to materialise all intermediate neighbours. As an optimisation, we instead only limit the final neighbourhood size, which greatly improves the time and resource usage. Our query is shown in Figure~\ref{fig:sampling-query}.

\begin{figure}
 \centering
 \begin{lstlisting}[language=cypher]
    MATCH (node_0:$NODE_TYPE)
    WHERE node_0.id IN $SEED_NODES
    OPTIONAL MATCH (node_0)-[rel_1:$REL_TYPE]->(node_1:$NODE_TYPE)-[rel_2:$REL_TYPE]->(node_2:$NODE_TYPE)
    WITH node_0, node_1, node_2
    ORDER BY rand()
    LIMIT $MAX_NEIGHBOURS
    RETURN 
      node_0.id as src_id, 
      node_1.id, node_1.features,
      node_2.id, node_2.features;
 \end{lstlisting}
 \caption{A Cypher query template that samples two-hop neighbourhoods of given seed nodes. Features of sampled nodes are returned in the same step.}
 \label{fig:sampling-query}
\end{figure}

\paragraph{Feature retrieval}
Since we no longer store the entire graph in-memory, we also offload feature storage to the database. One option is to use separate storage for both the features (such as vector databases\cite{pinecone}) and the graph. However, the storage formats of most graph DBs and their query engines allow efficient retrieval of features when retrieving the node IDs since node attributes are co-located with node IDs in storage\cite{neo4j}. This allows us to use one graph DB as the storage, selecting relevant attributes as shown in Figure\ref{fig:sampling-query}. The engineering advantage of not needing to set up a separate feature store and maintaining cross-DB consistency is non-negligible.

\paragraph{Distributed training}
Distributed training is enabled by the graph partitioning methods introduced in Section~\ref{subsec:background-training-architecture}. Existing graph partitioning methods are often slow and nondeterministic (which affects reproducibility). Additionally, they require re-partitioning for any modification of the graph. While inter-process communication overhead is minimised by workload-balancing min-cut partitioning, such communication still happens, which introduces unpredictable latency or could become a failure point in a system. Communication free partitioning\cite{cofree-gnn} comes at the cost of additional halo nodes.

If we take a step back, this is one solution to the problem of \textit{how to store data that allows efficient concurrent reads}. This is exactly what is solved by graph databases that support horizontal scaling for OLAP transactions\cite{besta2023graph}. The partitioning solution is commonly referred to as graph sharding\cite{powerlyra,chen2021g}, while simply duplicating copies of the data is another viable option.

\begin{figure}
    \centering
    \includegraphics[width=0.9\linewidth]{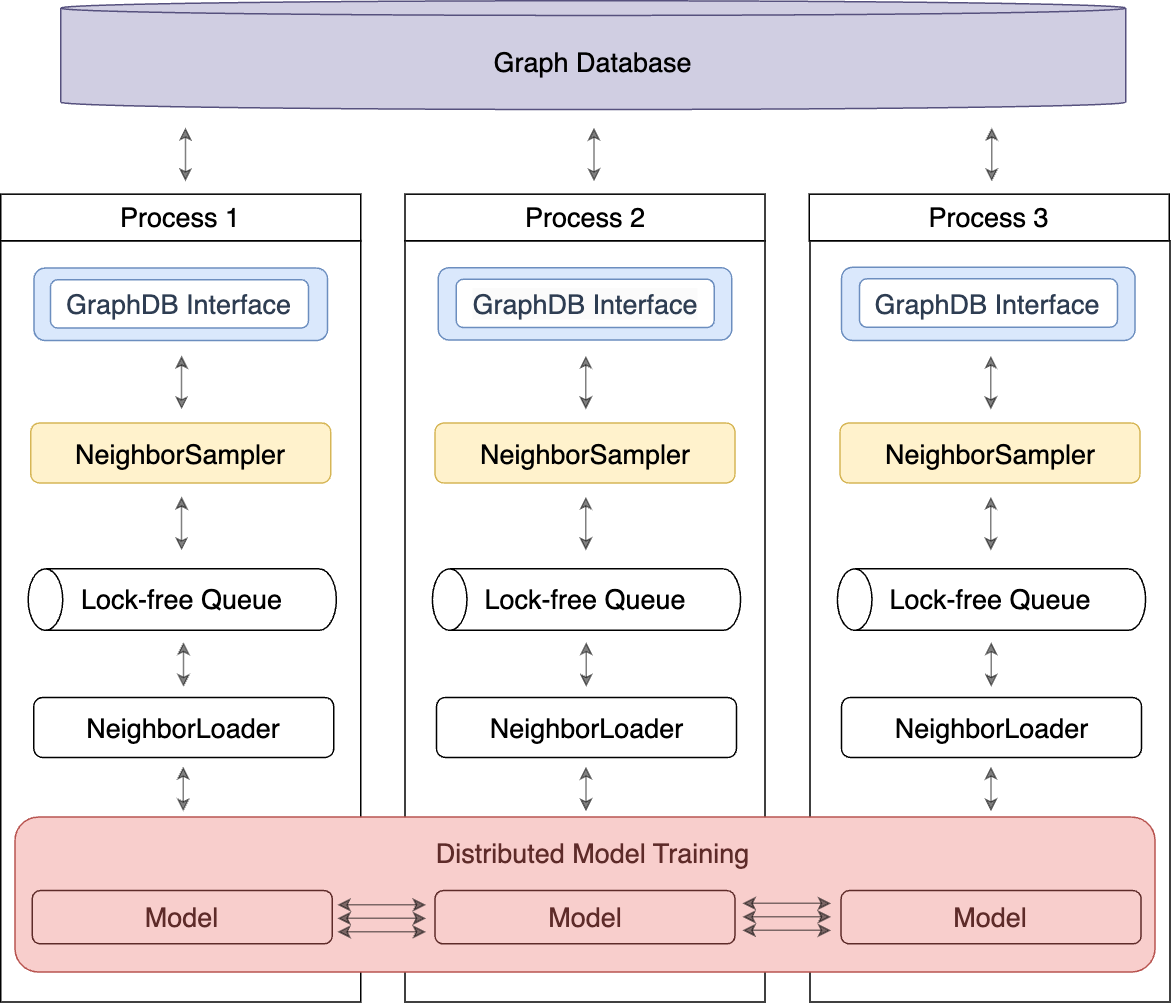}
    \caption{Our distributed training architecture. One graph database acts as a central graph and feature store. Multiple training processes can concurrently sample from the DB with no overhead. Our architecture closely mimics that of PyTorch Distributed Data Parallel\cite{dist-PyTorch} and applies to any other similar setup.}
    \label{fig:distributed-training}
\end{figure}

Since we have formulated sampling operations as a special type of read transaction, our method enables distributed training almost for free. One simply sets up the distributed training architecture as usual and configures each process to sample from the same database, as illustrated in Figure~\ref{fig:distributed-training}.

\section{Experiments}
We validate the applicability and scalability of our approach. We use ogbn-papers100M and ogbn-products \cite{hu2020open}, two real-world large graphs with different topology. We use Neo4j as our graph database. We refer to Appendix \ref{appendix:configuration} for our database, hardware and other training configurations. A detailed ablation study of performance with respect to different software stacks and database vendors is out of the scope of this work.  We use PyG and adapt its \texttt{GraphStore} and \texttt{FeatureStore} to be interfaces for Neo4j and implement a custom \texttt{Neo4jGraphSampler}. We benchmark our approach for both single-machine and distributed training setups. Our code is available on GitHub.\footnote{https://github.com/DmytroLopushanskyy/ScalingGNNs}

\subsection{Analysis of the query plan}
Formulating sampling as a special type of retrieval query is a central component of our approach. We verify that our query is indeed correct. We utilise the profiling tool provided by the query engine (keyword \texttt{PROFILE}) to examine the execution details of all the operators involved.

As shown in Table~\ref{tab:query-planner}, the correct number of initial and final rows are produced, and at each hop, we are able to verify the number of total neighbours (88, 18524). Two one-hop traversals (steps 8 and 6) are performed with filters applied optimally. An alternative execution plan could be first finding the cross product of all possible \texttt{node\_0, node\_1, node\_2} and then applying filtering by edges and labels, which is clearly suboptimal. 

The full execution plan also includes additional measurements such \texttt{Page Cache Hits/Misses}, \texttt{Time (ms)} for each operator, which we leave out here. Examining the execution plan allows one to understand the optimality of the query, benchmark detailed runtime behaviour, and potentially find improvements in either the written query or the query engine implementation.

\begin{table}
    \centering
    \caption{An example query execution plan for Figure~\ref{fig:sampling-query} with a specific set of 10 \texttt{\$SEED\_NODES} and 100 \texttt{\$MAX\_NEIGHBOURS}. \texttt{Operator} are executed from the bottom up. \texttt{Details} represent the exact execution parameters. \texttt{Estimated Rows} represent expected rows produced. \texttt{Rows} represent actual rows produced. \texttt{DB Hits} measure the amount of work by the storage engine. Total database access is 38570, total allocated memory is 19120 bytes.}
    \label{tab:query-planner}
    \ttfamily 
    \resizebox{\textwidth}{!}{ 
    \begin{tabular}{|l|l|l|l|l|l|}
        \toprule
        \multirow{2}{*}{Operator}         & \multirow{2}{*}{Id}  & \multirow{2}{*}{Details}                   & \multirow{2}{*}{\makecell{Estimated \\ Rows}} & \multirow{2}{*}{Rows}  & \multirow{2}{*}{DB Hits} \\
        & & & & & \\
        \midrule
        +ProduceResults  & 0   & node\_0, node\_1, node\_2 & 100              & 100     & 1102  \\
        \cmidrule(lr){2-6}
        +Top             & 1   & rand() ASC LIMIT 100    & 100              & 100     & 0     \\
        \cmidrule(lr){2-6}
        +Projection      & 2   & rand() AS `rand()`      & 8381             & 18524   & 0     \\
        \cmidrule(lr){2-6}
        +Apply           & 3   &                         & 8381             & 18524   & 0     \\
        \cmidrule(lr){2-6}
        |+Optional        & 4   & node\_0                  & 8381             & 18524   & 0     \\
        \cmidrule(lr){2-6}
        |+Filter          & 5   & NOT anon\_1 = anon\_0 AND node\_2:PAPER   & 8381    & 18524   & 18524\\
        \cmidrule(lr){2-6}
        |+Expand(All)     & 6   & (node\_1)-[anon\_1:CITES]-(node\_2)       & 8466    & 18612   & 18738\\
        \cmidrule(lr){2-6}
        |+Filter          & 7   & node\_1:PAPER           & 291                 & 88   & 88 \\
        \cmidrule(lr){2-6}
        |+Expand(All)     & 8   & (node\_0)-[anon\_0:CITES]-(node\_1)       & 291  & 88   & 98 \\
        \cmidrule(lr){2-6}
        +Argument        & 9   & node\_0                  & 10             & 10   & 0 \\
        \cmidrule(lr){2-6}
        \multirow{2}{*}{+NodeIndexSeek}   & \multirow{2}{*}{10}   & \multirow{2}{*}{\makecell{TEXT INDEX node\_0:PAPER(id) \\ WHERE id IN \$autolist\_0, cache[node\_0.id]}}  & \multirow{2}{*}{10} & \multirow{2}{*}{10} & \multirow{2}{*}{20} \\
         & & & & & \\
        \bottomrule
    \end{tabular}
    }
\end{table}

\subsection{Single machine training}
We now study the runtime characteristics of our setup for single-machine training. Our setup should support training extremely large graphs (ogbn-papers100M) on tiny machines with low RAM usage, since at any point in time, the total RAM required is no larger than representing the 2-hop sampled neighbourhood of batch-size number of nodes, plus some model weights. At the same time, as we equip the training machine with more hardware resources, we expect training speed to improve. Since the absolute wall-clock time is only meaningful to the specific hardware, we compare our setup using a Neo4j database with a common training setup using no database and with an embedded graph database \cite{kuzudb}. In order to benchmark the runtime trade-off that results from using a DB, we use one CPU only. We train GraphSAGE represented by Equation~\ref{eq:graphsage} for node classification.

\begin{table}
	\centering
	\caption{Single-machine training. Batch Time measures the average time per batch. Sampled Nodes record average sampled nodes per batch and, similarly, Sampled Edges.}
	\label{tab:ogbn-papers-result}
	\begin{tabular}{cccccc}
		\toprule
            Backend & RAM & Batch Time & Sampled Nodes & Sampled Edges & Batch Size \\
            \midrule
            No Backend & 256 GB & 7.6s & 23 898 & 26 653 & 1024 \\
		No Backend & 128 GB & \multicolumn{3}{c}{Out-Of-Memory} & 1024 \\
            Embedded & 128 GB & 11s & 31 768 & 32 670 & 1024 \\
            Embedded & 60 GB & \multicolumn{3}{c}{Out-Of-Memory} & 1024 \\
            Disk-based & 256 GB & 75s & 24 138 & 48 907 & 1024 \\
            \textbf{Disk-based} & \textbf{128 GB} & \textbf{76s} & \textbf{24 081} & \textbf{48 899} & \textbf{1024} \\
            Disk-based & 12 GB & 108s & 23 979 & 48 905 & 1024 \\
            \textbf{Disk-based} & \textbf{8 GB} & \textbf{126s} & \textbf{23 678} & \textbf{48 909} & \textbf{512} \\
		\bottomrule
	\end{tabular}
\end{table}

As shown in Table~\ref{tab:ogbn-papers-result}, our training works even when there is only 8GB of RAM available, using a batch size of 512. As we increase RAM usage, the average processing time decreases, or one could equivalently increase batch sizes. Therefore our setup enables a smooth trade-off between hardware resources and training time for an arbitrary scale. For example, our approach should enable graph ML workloads for extremely large graphs, such as Graph500-scale34 \cite{graph500} with more than 17 billion nodes and 274 billion edges, commonly used for HPC graph analytics but have not been achieved within the graph ML community.

We also examine what happens when there is an abundance of memory available that allows common single-machine training. As expected, the no-backend setup is on par with an embedded DB, both being significantly faster than using a separate graph DB. This suggests the IO processing and converting Cypher output into PyTorch tensor (an ETL step, which is a non-zero-copy operation) becomes the new bottleneck.

\subsection{Distributed training}
We implement Figure~\ref{fig:distributed-training} and study the runtime characteristics of distributed training using a graph DB.

Since we already understand the trade-off between memory usage and processing power, we fix all experiments with 128GB RAM. A basic feature of any database is to support many concurrent reads. While each single query's execution is single-threaded, we expect a large number of concurrent queries to utilise a large number of threads to saturate available CPUs.

To make sure we are benchmarking only the impact of introducing the database, but not the orthogonal concern of unpredictable model synchronisation and network latency, we use a single machine with 16 CPUs and multiple separate training processes to simulate the distributed setup. 

\begin{table}
	\centering
	\caption{Distributed training. Partitions represent the number of training processes.}
	\label{tab:distributed-results}
	\begin{tabular}{cccccc}
		\toprule
		Backend & Batch Size & Partitions & Epoch Time & Sampled Nodes & Sampled Edges\\
		\midrule
            \multirow{2}{*}{No Backend} & \multirow{2}{*}{1024} & 2 & 1061s & 406 372 & 713 840 \\ \cmidrule(lr){3-6}
            & & \textbf{8} & \textbf{461s} & 410 237 & 716 912 \\
            \midrule
            \multirow{9}{*}{Graph DB} & \multirow{4}{*}{512} & \textbf{2} & \textbf{1 841s} & 9 024 & 46 073 \\ \cmidrule(lr){3-6}
            &  & \textbf{8} & \textbf{564s} & 7 916 & 46 061 \\ \cmidrule(lr){3-6}
            &  & 12 & 597s & 7 403 & 46 065 \\ \cmidrule(lr){3-6}
            &  & 16 & 629s & 9 037 & 46 051 \\ \cmidrule(lr){2-6}
            \cmidrule(lr){2-6}
            & \multirow{3}{*}{1024} & \textbf{2} & \textbf{2 094s} & 7 027 & 92 107 \\ \cmidrule(lr){3-6}
            &  & \textbf{8} & \textbf{600s} & 18 215 & 92 042 \\ \cmidrule(lr){3-6}
            &  & 16 & 590s & 15 563 & 92 108 \\ \cmidrule(lr){2-6}
            \cmidrule(lr){2-6}
            & \multirow{2}{*}{8192} & 2 & 6 676s & 104 425 & 737 118 \\ \cmidrule(lr){3-6}
            &  & 8 & 1 702s & 112 082 & 736 857 \\
		\bottomrule
	\end{tabular}
\end{table}

As shown in Table~\ref{tab:distributed-results}, as we increase the number of parallel training processes from 2 to 8, training speed improves linearly by about 4 times. The optimal epoch time is achieved with 8 processes. Increasing the number of training processes no longer improves training speed due to the fact that all CPUs are now saturated. 

We additionally check the correctness of our setup by examining the empirical distribution of sampled nodes in Appendix~\ref{appendix:node-distribution} and verify that it aligns with the degrees of sampled nodes. Our model gives an average of 58\% accuracy after 10 epochs without hyperparameter and batch tuning for model weight synchronisation. Industry standard model architecture, hyperparameter and DDP configuration tuning should be used in real-world deployments to reach state-of-the-art scores.

\section{Conclusion}
In this paper, we introduce a new way of training GNNs by using graph DB as an integral component in the training pipeline. We formulate common data access operations in GNN training pipelines as retrievals in graph DB. Concretely, initial full-graph loading is removed, and only metadata is loaded, in addition to multi-hop neighbourhood sampling with feature retrieval being executed as one query. 

Our experiments validate that minimal memory is required to train arbitrarily large graphs. At the same time, there is a smooth scaling curve of training speed with respect to hardware resources. Our setup also seamlessly applies to distributed training, where similar memory advantage and scaling behaviour are observed. The separate pre-processing step of graph partitioning is not required since this is exactly what is offered by database horizontal scaling.

Most importantly, our approach leverages the techniques and tooling from graph DB to the benefit of graph ML and lays a foundation for future developments at the intersection of these two fields.

\section{Future Work}
There is a variety of future work that builds upon our approach, and we outline several promising follow-up directions.

\paragraph{Engineering optimisations}
Since this is the first work that tightly couples a GNN training process with a graph DB, our setup is focused on verifying the method's applicability and benchmarking the trade-offs. For real-world deployments, hardware accelerators such as GPUs/TPUs can be leveraged to further speed up the training process in a standard way. The common bottleneck of workload imbalanced between sampling and tensor operations could be addressed with scheduling methods similar to \cite{bytegnn, g3}. In single-machine training with an embedded graph DB, memory I/O could be optimised by using shared memory layouts between GNN processing and DB query engine. Database indexes could be created in a way that matches the retrieval patterns of GNN workloads.

\paragraph{Other sampling methods}
While our query in Figure~\ref{fig:sampling-query} can be modified to support common node-wise sampling schemes, it is unclear whether more exotic sampling methods\cite{liu2021sampling} can be formulated as queries. For example, there exist many heterogeneous GNNs \cite{fu2020magnn, han, metapath2vec, hetgnn}, all of which are based on neighbourhood sampling with respect to some metapaths. A metapath is essentially a node-label and edge-type sequence, such as 
\begin{equation*}
    \text{AUTHOR} \xrightarrow{\text{WRITES}} \text{PAPER} \xrightarrow{\text{CITES}} \text{PAPER} \xleftarrow{\text{WRITES}} \text{AUTHOR}
\end{equation*}

LPGs can naturally model any heterogeneous graphs. It is clear that such metapaths can be easily defined as a query, such as 
\begin{center}
    \texttt{MATCH (n1:AUTHOR)-[e1:WRITES]->(n2:PAPER)-[e2:CITES]->} \\
    \texttt{(n3:PAPER)<-[e3:WRITES]-(n4:AUTHOR)}
\end{center}

Moreover, such heterogeneous long multi-hop queries are well supported by OLAP graph databases.

Both the formal study of which sampling methods can be modelled within query languages and the empirical experiments of which method sees the biggest gain from using graph DBs are interesting future directions.

\paragraph{Further integration with graph DBs}
Our query in Figure~\ref{fig:sampling-query} returns the multiset in Equation~\ref{eq:graphsage}. The immediate next step in the training pipeline is to apply mean aggregation, which is also a basic aggregation operator in the query. Therefore, one could attempt to implement more parts of Equation~\ref{eq:graphsage}, or perhaps all of it, within the database. The natural question is which particular instance of Equation~\ref{eq:mpnn} can not be implemented, and in which cases do we observe practical advantages.

\paragraph{Inference}
Typical inference workflow mimics the forward pass of the training setup. As a result, classifying a single new node requires both recreating the graph and a full forward pass. It seems our setup could be extended to support extremely resource-efficient and fast inference of new data with minimal modification.

\section*{Acknowledgements}

The authors thank Michael Benedikt for the valuable comments and discussions on this work.

\bibliographystyle{unsrtnat}
\bibliography{reference}

\appendix
\section{Cypher query for retrieving edge metadata}\label{appendix:edge-metadata}
Figure~\ref{fig:edge-metadata} shows an example Cypher query that returns all relevant edge metadata.
\begin{figure}[h]
 \centering
 \begin{lstlisting}[language=cypher]
    MATCH (a)-[r]->(b)
    WITH DISTINCT type(r) AS EdgeType, labels(a)[0] AS SourceType, labels(b)[0] AS TargetType, r
    WITH EdgeType, SourceType, TargetType, collect(distinct keys(r)) AS AllKeys, count(r) AS edge_count
    RETURN EdgeType, SourceType, TargetType, reduce(s = [], k IN AllKeys | s + k) AS UniqueKeys, edge_count
 \end{lstlisting}
 \caption{A Cypher query that returns all relevant metadata for all edges.}
 \label{fig:edge-metadata}
\end{figure}

\section{Single-hop sampling query}\label{appendix:one-hop-sampling}
Figure~\ref{fig:one-hop-sample} shows a Cypher query that performs one-hop neighbourhood sampling. The nodes returned can be used as seed nodes for subsequent sampling on the next hop.
\begin{figure}[h]
 \centering
 \begin{lstlisting}[language=cypher]
    MATCH (node_src:PAPER)-[rel:CITES]-> (node_dst:PAPER)
    WHERE id(node_src) IN $SEED_NODES
    RETURN id(node_dst), rand() as r
    ORDER BY r
    LIMIT $MAX_NEIGHBOURS
 \end{lstlisting}
 \caption{A Cypher query that samples one-hop neighbourhood for given seed nodes.}
 \label{fig:one-hop-sample}
\end{figure}

\section{Database and training configuration} \label{appendix:configuration}
We use Neo4j Community Edition version 5.21. We configure \texttt{dbms.memory.heap.initial\_size = 32GB}, \texttt{dbms.memory.heap.max\_size = 32GB} and \texttt{dbms.memory.pagecache.size=32GB}. We use Kuzu version 0.5.2. All other Neo4j and KuzuDB configurations are kept as default. We use PyTorch 2.3.0 and PyG 2.5.0. When sampling two-hops we use $12\times12 = 144$ number of neighbours.

\section{Empirical distribution of sampled nodes} \label{appendix:node-distribution}
\begin{figure}
    \centering
    \includegraphics[width=1.0\linewidth]{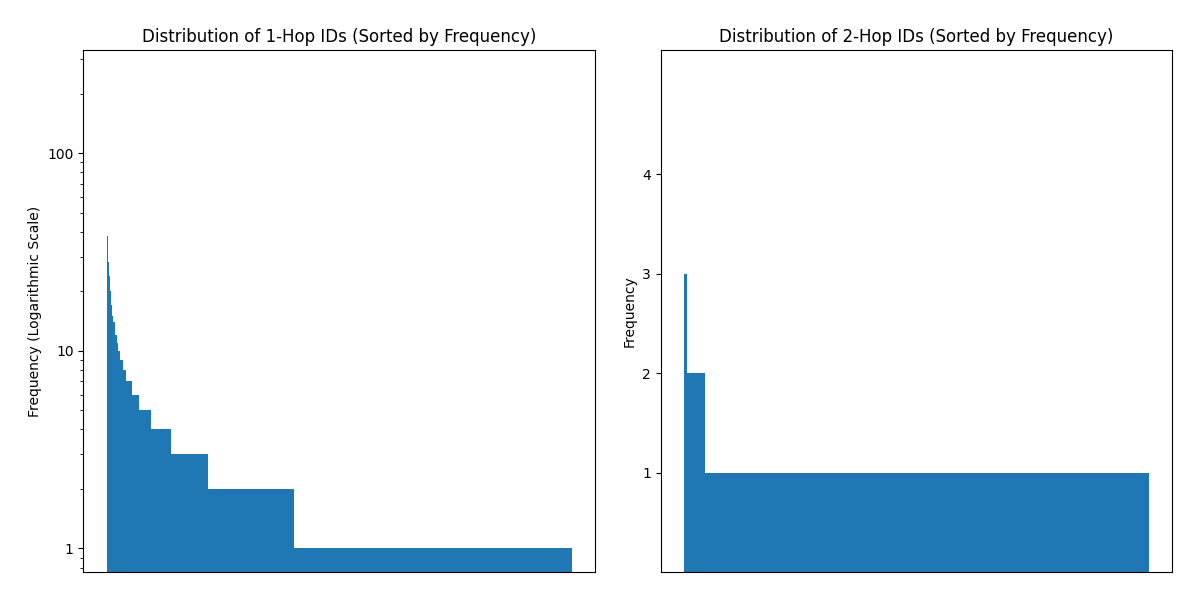}
    \caption{Empirical distribution of sampled nodes using query template Figure~\ref{fig:sampling-query}.}
    \label{fig:node-distribution}
\end{figure}

Figure~\ref{fig:node-distribution} shows the plots of sampled nodes. We run the query in Figure~\ref{fig:sampling-query} 100 times on ogbn-products, with \$SEED\_NODES of 100 nodes from id 100000 to 100099. The empirical distribution of frequency of sampled nodes aligns with actual degree distribution, with a few high-degree nodes sampled frequently and the majority of nodes sampled once.

\end{document}